\documentclass[11pt,a4paper]{article}
\usepackage[hyperref]{emnlp2018}
\usepackage{times}
\usepackage{latexsym}
\usepackage{graphicx}

\usepackage{tikz}

\usepackage{url}

\aclfinalcopy 

\title{A Multilingual Information Extraction Pipeline\\ for Investigative Journalism}

\author{Gregor Wiedemann \qquad Seid Muhie Yimam \qquad Chris Biemann \\
Language Technology Group\\ 
Department of Informatics\\ 
Universit\"{a}t Hamburg, Germany\\
 {\tt \{gwiedemann, yimam, biemann\}@informatik.uni-hamburg.de}
}
\date{}

\begin{document}
\maketitle
\begin{abstract}
We introduce an advanced information extraction pipeline to automatically process very large collections of unstructured textual data for the purpose of investigative journalism.
The pipeline serves as a new input processor for the upcoming major release of our \emph{New/s/leak 2.0} software,  which we develop in cooperation with a large German news organization. 
The use case is that journalists receive a large collection of files up to several Gigabytes containing unknown contents. Collections may originate either from official disclosures of documents, e.g. Freedom of Information Act requests, or unofficial data leaks. 
Our software prepares a visually-aided exploration of the collection to quickly learn about potential stories contained in the data.
It is based on the automatic extraction of entities and their co-occurrence in documents.
In contrast to comparable projects, we focus on the following three major requirements particularly serving the use case of investigative journalism in cross-border collaborations: 1) composition of multiple state-of-the-art NLP tools for entity extraction, 2) support of multi-lingual document sets up to 40 languages, 3) fast and easy-to-use extraction of full-text, metadata and entities from various file formats.
\end{abstract}

\section{Support Investigative Journalism}

Journalists usually build up their stories around entities of interest such as persons, organizations, companies, events, and locations in combination with the complex relations they have. This is especially true for investigative journalism which, in the digital age, more and more is confronted to find such relations between entities in large, unstructured and heterogeneous data sources. 

Usually, this data is buried in unstructured texts, for instance from scanned and OCR-ed documents, letter correspondences, emails or protocols. Sources typically range from 1) official disclosures of administrative and business documents, 2) court-ordered revelation of internal communication, 3) answers to requests based on Freedom of Information (FoI) acts, and 4) unofficial leaks of confidential information. Well-known examples of such disclosed or leaked datasets are the \emph{Enron} email dataset \cite{Keila.2005b} or the \emph{Panama Papers} \cite{ODonovan.2016}.

To support investigative journalism in their work, we have developed \emph{New/s/leak} \cite{yimam-EtAl:2016:P16-4}, a software implemented by experts from natural language processing and visualization in computer science in cooperation with journalists from \emph{Der Spiegel}, a large German news organization. Due to its successful application in the investigative research as well as continued feedback from academia, we further extend the functionality of \emph{New/s/leak}, which now incorporates better pre-processing, information extraction and deployment features. 
The new version \emph{New/s/leak 2.0} serves four central requirements that have not been addressed by the first version or other existing solutions for investigative and forensic text analysis:

    \noindent \textbf{Improved NLP processing}:  We use stable and robust state-of-the-art natural language processing (NLP) to automatically extract valuable information for journalistic research. Our pipeline combines extraction of temporal entities, named entities, key-terms, regular expression patterns (e.g. URLs, emails, phone numbers) and user-defined dictionaries.\\
    \noindent \textbf{Multilingualism}: Many tools only work for English documents or a few other `big languages'. In the new version, our tool allows for automatic language detection and information extraction in 40 different languages. Support of multilingual collections and documents is specifically useful to foster cross-country collaboration in journalism. \\
    \noindent \textbf{Multiple file formats}: Extracting text and metadata from various file formats can be a daunting task, especially in journalism where time is a very scarce resource. In our architecture, we include a powerful data wrangling software to automatize this process as much as possible. We further put emphasis on scalability in our pipeline to be able to process very large datasets. For easy deployment, \emph{New/s/leak 2.0} is distributed as a Docker setup.\\
    \noindent \textbf{Keyword graphs}: We have implemented keyword network graphs, which is build based on the set of keywords representing the current document selection. The keyword network enables to further improve the investigation process by displaying entity networks related to the keywords.

\section{Related Work}
\label{related}
There are already a handful of commercial and open-source software products to support investigative journalism. Many of the existing tools such as OpenRefine\footnote{\url{http://openrefine.org}}, Datawrapper\footnote{\url{https://www.datawrapper.de}}, Tabula\footnote{\url{http://tabula.technology}}, or Sisense\footnote{\url{https://www.sisense.com}} focus solemnly on structured data and most of them are not freely available. 
For unstructured text data, there are costly products for forensic text analysis such as Intella\footnote{\url{https://www.vound-software.com}}. Targeted user groups are national intelligence agencies. For smaller publishing houses, acquiring a license for those products is simply not possible.
Since we also follow the main idea of openness and freedom of information, we concentrate on other open-source products to compare our software to.

DocumentCloud\footnote{\url{https://www.documentcloud.org}} is an open-source tool specifically designed for journalists to analyze, annotate and publish findings from textual data. In addition to full-text search, it offers named entity recognition (NER) based on OpenCalais\footnote{\url{http://www.opencalais.com}} for person and location names. In addition to automatic NER for multiple languages, our pipeline supports the identification of keyterms as well as temporal and user-defined entities.

Overview \cite{Brehmer2014} is another open-source application developed by computer scientists in collaboration with journalists to support investigative journalism. The application supports import of PDF, MS Office, and HTML documents, document clustering based on topic similarity, a simple location entity detection, full-text search, and document tagging. Since this tool is already mature and has successfully been used in a number of published news stories, we adapted some of its most useful features such as document tagging, full-text search and a keyword-in-context (KWIC) view for search hits. 

The Jigsaw visual analytics \cite{Carsten2014} system is a third tool that supports analyzing and understanding of textual documents. The Jigsaw system focuses on the extraction of entities using Gate tool suite for NLP \cite{Cunningham.2013}. Hence, support for multiple languages is somewhat limited. It also lacks sophisticated data import mechanisms.


The new version of \emph{New/s/leak} was built targeting these drawbacks and challenges. With \emph{New/s/leak 2.0} we aim to support the journalist throughout the entire process of collaboratively analyzing large, complex and heterogeneous document collections: data cleaning and formatting, metadata extraction, information extraction, interactive filtering, visualization, close reading and tagging, and providing provenance information.

\section{Architecture}
\label{arch}
\begin{figure*}
    \centering
	\includegraphics[width=0.8\linewidth,trim={0 0 0 0},clip]{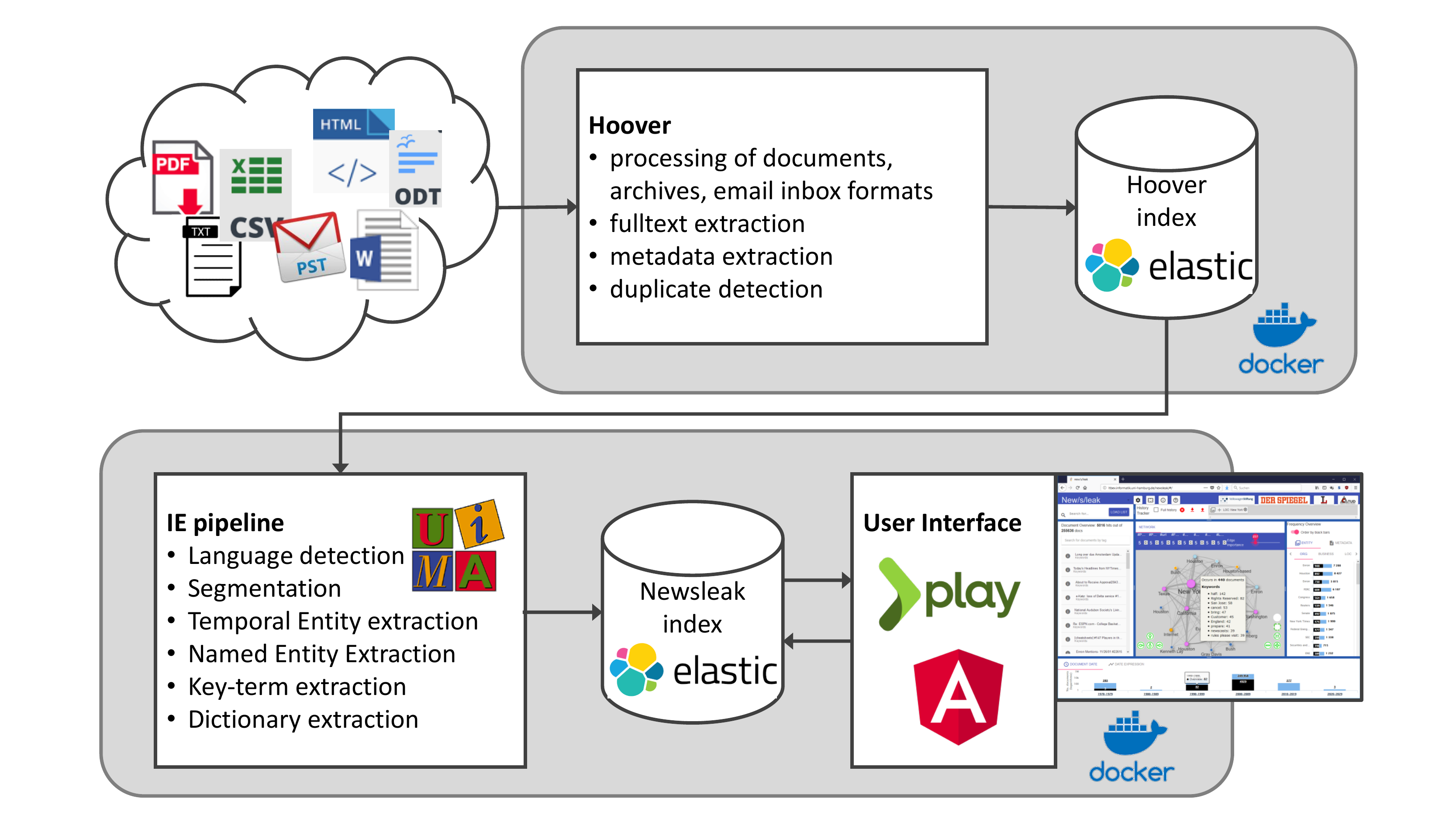}
	\caption{Architecture of \emph{New/s/leak 2.0}}
	\label{fig:arch} 
	\end{figure*}

Figure \ref{fig:arch} shows the overall architecture of \emph{New/s/leak}. In order to allow users to analyze a wide range of document types, our system includes a document processing pipeline, which extracts text and metadata from a variety of document types into a unified representation. On this unified text representation, a number of NLP pre-processing tasks are performed as a UIMA pipeline \cite{Ferrucci.2004b}, e.g. automatic identification of the document language, segmentation into paragraph, sentence and token units, and extraction of named entities, keywords and metadata. ElasticSearch is used to store the processed data and create aggregation queries for different entity types to generate network graphs. The user interface is implemented with a RESTful web service based on the Scala Play framework in combination with an AngularJS browser app to present information to the journalists. Visualizations are realized with D3 \cite{Bostock.2011}. 

In order to enable a seamless deployment of the tool by journalists with limited technical skills, we have integrated all of the required components of the architecture into a Docker\footnote{\url{https://www.docker.com}} setup. Via docker-compose, a software to orchestrate Docker containers for complex architectures, end-users can download and run locally a preconfigured version of \emph{New/s/leak} with one single command. Being able to process data locally and even without any connection to the internet is a vital prerequisite for journalists when they work with sensitive data. All necessary source code and installation instructions can be found on our Github project page.\footnote{\url{https://uhh-lt.github.io/newsleak-frontend}}

\section{Data Wrangling}
\label{sec:data_wrangling}

Extracting text and metadata from various formats into a format readable by a specific analysis tool can be a tedious task. In an investigative journalism scenario, it can even be a deal breaker since time is an especially scarce resource and file format conversion might not be a task journalists are well trained in. 
To offer access to as many file formats as possible in \emph{New/s/leak}, we opted for a close integration with \emph{Hoover},\footnote{\url{https://hoover.github.io}} a set of open-source tools for text extraction and search in large text collections. Hoover is developed by the European Investigative Collaborations (EIC) network with a special focus on large data leaks and heterogeneous datasets. It can extract data from various text file formats such as txt, html, docx, pdf, zip, tar, pst, mbox, eml, etc. The text is extracted along with metadata from files (e.g. file name, creation date, file hash) and header information (e.g. subject, sender, and receiver). In the case of emails, attachments are processed automatically, too.

\emph{New/s/leak} connects directly to Hoover's index to read full-texts and metadata for its own information extraction pipeline. Through this close integration with Hoover, \emph{New/s/leak} can offer information extraction to a wide variety of data formats. In many cases, this drastically limits or even completely eliminates the amount of work needed to clean and preprocess large datasets beforehand.

\section{Multilingual Information Extraction}
\label{sec:multilang_ner}

The core functionality of \emph{New/s/leak} is the automatic extraction of various kinds of entities from text to facilitate the exploration and sense-making process from large collections. Since a lot of steps in this process involve language-dependent resources, we put an emphasis on the work to support as many languages as possible.

\subsection{Preprocessing}

Information extraction in \emph{New/s/leak} is implemented as a configurable UIMA pipeline \cite{Ferrucci.2004b}. Text documents and metadata from a Hoover collection (see Section~\ref{sec:data_wrangling}) are read in parallelized manner and put through a chain of annotators. In a final step of the chain, results from annotation processes are indexed in an ElasticSearch index for later retrieval and visualization.

First, we identify the language of each document. Alternatively, language can also be determined on a paragraph level to support multi-language documents, which can occur quite often, for instance in email leaks or bilingual contracts. Second, we separate sentences and tokens in each text. To guarantee compatibility with various Unicode scripts in different languages, we rely on the ICU4J library\footnote{\url{http://icu-project.org/apiref/icu4j}} for this task. ICU4J provides locale-specific sentence and word boundary detection relying on a simple rule-based approach. While the quality of the segmentation and tokenization results might be better when using specifically trained segmentation models, the advantage of the rule-based approach in ICU4J is that it works robustly not only for many languages but also for noisy data, which we expect to be abundant in real-life datasets.

\subsection{Dictionaries and RE-patterns}

In many cases, journalists follow some hypothesis to test for their investigative work. Such a proceeding can involve looking for mentions of already known terms or specific entities in the data. This can be realized by lists of dictionaries provided to the initial information extraction process. \emph{New/s/leak} annotates every mention of a dictionary term with the respective list type. Dictionaries can be defined in a language-specific fashion, but also applied across documents of all languages in the corpus. Extracted dictionary entities are displayed along with extracted named entities in the visualization.

In addition to self-defined dictionaries, we annotate email addresses, telephone numbers, and URLs with regular expression patterns. This is useful, especially for email leaks to reveal communication networks of persons and filter for specific email account related content.

\subsection{Temporal Expressions}

Tracking documents across the time of their creation or by temporal events they mention can provide valuable information during investigative research.
Unfortunately, many document sets (e.g. collections of scanned pages) do not come with a specific document creation date as structured metadata. To offer a temporal selection of contents to the user, we extract mentions of temporal expressions. This is done by integrating the HeidelTime temporal tagger \cite{Strotgen.2015} in our UIMA workflow. HeidelTime provides automatically learned rules for temporal tagging in more than 200 languages. Extracted timestamps can be used to select and filter documents.

\subsection{Named Entity Recognition}

We automatically extract person, organization and location names from all documents to allow for an entity-centric exploration of the data collection. Named entity recognition is done using the \emph{polyglot-NER} library \cite{AlRfou.2015}. Polyglot-NER contains sequence classification for named entities based on weakly annotated training data automatically composed from Wikipedia\footnote{\url{https://wikipedia.org}} and Freebase\footnote{\url{https://developers.google.com/freebase}}. Relying on the automatic composition of training data allows polyglot-NER to provide pre-trained models for 40 languages.\footnote{A list of the 40 languages covered by Polyglot-NER can be found at \url{https://tinyurl.com/yaju7bf7}}


\subsection{Keyterm Extraction}

To further summarize document contents in addition to named entities, we automatically extract keyterms and phrases from documents. For this, we implemented a keyterm extraction library for the 40 languages also supported in the previous step.\footnote{\url{https://github.com/uhh-lt/lt-keyterms}} Our approach is based on a statistical comparison of document contents with generic reference data. Reference data for each language is retrieved from the Leipzig Corpora Collection \cite{Goldhahn.2012}, which provides large representative corpora for language statistics. We employ log-likelihood significance as described in \cite{Rayson.2004} to measure the overuse of terms (i.e. keyterms) in our target documents compared to the generic reference data. Ongoing sequences of keyterms in target documents are concatenated to key phrases if they occur regularly in that exact same order. Regularity is determined with the Dice coefficient. This simple method allows to reliably extract multiword units such as ``stock market'' or ``machine learning'' in the documents. 
Since this method also extracts named entities if they occur significantly often in a document, there can be a substantial overlap between both types. To allow for a separate display of named entities and keywords, we filter keyterms if they already have been annotated as a named entity.
The remaining top keyterms are used to create a brief summary of each document for the user and to generate keyterm networks for document browsing.

\section{User Interface}
\label{entkey}

\begin{figure}
    \centering
	\includegraphics[width=\linewidth]{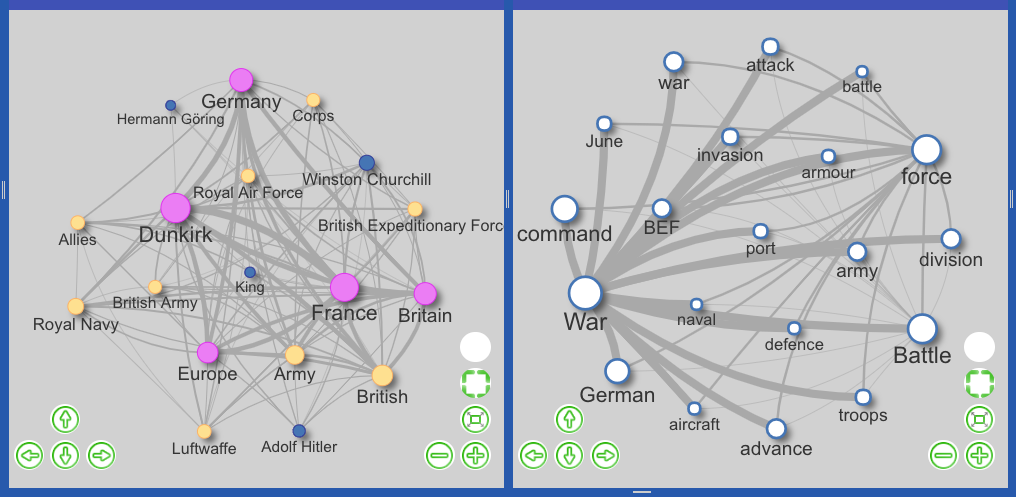}
	\caption{The entity and keyword graphs of \emph{New/s/leak} based on the WW2 collection (see Section~\ref{casestudy}). Networks are visualized based on the current document selection, which can be filtered by full-text search, entities or metadata. Visualization parameters such as the number of nodes per type or minimum edge strength can be set by the user. 
	Hovering over nodes and edges in one graph highlights information present in the respective another graph to show which entities and keywords frequently co-occur with each other in documents.}
	\label{fig:entkey} 
	\end{figure}

\paragraph{Browsing entity networks:}
Access to unstructured text collections via named entities is essential for journalistic investigations. To support this, we included two types of graph visualization, as it is shown in Figure \ref{fig:entkey}. The first graph, called entity network, displays entities in a current document selection as nodes and their joint occurrence as edges between nodes. Different node colors represent different types such as person, organization or location names. Furthermore, mentions of entities that are annotated based on dictionary lists are included in the entity network graph. The second graph, called keyword network, is build based on the set of keywords representing the current document selection. The keyword network also includes tags that can be attached to documents by journalists during work with the collection. 

\paragraph{Journalist in the loop:} In addition to the automatic annotation of entities and keyterms, we further enable journalists to: 1) annotate new entity types that are not in the system at all, 2) correct automatic annotations provided by the pipeline, e.g. to remove false positives or false entity type labels annotated by the NER process, 3) merge identical entities which have different forms (e.g. last names to full names, or spelling variants in different languages), and 4) label documents with user-defined terms called \emph{tags}. The tags are mainly used to annotate the document either for later reading or to share with collaborators.

\section{Case Study}
\label{casestudy}
To illustrate analysis capabilities of the new version of \emph{New/s/leak}, we present an exemplary case study at \url{https://ltdemos.informatik.uni-hamburg.de/newsleak/} (login with "user" and "password").
Since we refrain from publishing any confidential leak data, we created an artificial dataset from publicly available documents that share certain characteristics with the data from intended use cases in investigative journalism. It contains documents written in multiple languages, roughly centered on one topic and is full of references to entities.

Ca. 27.000 documents in our sample set are Wikipedia articles related to the topic of World War II. Articles were crawled from the encyclopedia in four languages (English:en, Spanish:es, Hungarian:hu, German:de) as a link network starting from the article "Second World War" in each respective language. Preprocessing and data extraction took around 75 minutes on a moderately fast server with 12 parallel CPU threads. 

\paragraph{Analysis:}
From a perspective of national history discourses and education, a certain common knowledge about WW2 can be expected. But, the topic becomes quickly a novel unexplored terrain for most people when it comes to aspects outside of the own region, e.g. the involvement of Asian powers. In our test case, we strive to fill gaps in our knowledge by identifying interesting details regarding this question. First, we start with a visualization of entities from the entire collection which highlights central actors of WW2 in general. In the list of extracted location entities, we can filter for ca. 2,000 articles referencing to Asia (en, es), \'{A}zsia (hu) or Asien (de). In this subselection, we find most references to China as a political power of the region followed by India and Japan. Further refinement of the collection by references to China highlights a central person name in the network, Chiang Kai-shek, who raises our interest. To find out more, we start the filter process all over again, subselecting all articles referencing this name. The resulting entity network reveals a close connection to the organization Kuomintang (KMT). Filtering for this organization, too, we can quickly identify articles centrally referencing to both entities by looking at their titles and extracted keywords. From the corresponding keyterm network and a KWIC view into the article full-texts, we learn that KMT is the national party of China and Kai-Shek as their leader ruled the country during the period of WW2. A second central actor, Mao Zedong, is strongly connected with both, KMT and Chiang Kai-shek in our entity network. From articles also prominently referencing Zedong, we learn from sections highlighting both person names that Kai-shek and Zedong, also a member of KMT and later leader of the Chinese Communists, shared a complicated relationship. By filtering for both names, we can now explore the nature of this relationship in more detail and compare its display across the four languages in our dataset.\footnote{A video of the proceeding can be found at: \url{http://youtu.be/96f_4Wm5BoU}}

\section{Discussion and Future Work}

In this paper, we introduced the completely renewed information extraction pipeline of \emph{New/s/leak 2.0}, an open-source software to support investigative journalism. 
As major requirements based on prior experiences, we identified the automatic annotation of various entity types in very large, multi-lingual document sets contained in heterogeneous file formats.
Our solution involves a combination of powerful NLP libraries for temporal and named entities, own developments for keyterm and pattern extraction, and a powerful data wrangling suite for text and metadata extraction.
The pipeline is capable to process information extraction in 40 languages.

\emph{New/s/leak} has been in use successfully at the German news organization \textit{Der Spiegel}. It recently has also been introduced as an open-source tool to the community of investigative journalists at respective conferences. We expect to collect more user feedback and experiences from case studies in the near future to further improve the software.

As a new main feature, we plan to extend the information extraction pipeline for user-defined categories into the direction of adaptive and active machine learning approaches. Currently, while reading the full-texts, users can manually annotate new entity types in the text or tag the entire documents. In combination with an adaptive and active learning approach, users will be able to train automatic tagging of documents and extraction of information while working with the data in the user interface.

\bibliographystyle{acl_natbib_nourl}
\bibliography{multinewsleak}

\end{document}